\pdfoutput=1 

\documentclass[conference]{IEEEtran}
\IEEEoverridecommandlockouts
\usepackage{cite}
\usepackage{amsmath,amssymb,amsfonts}
\usepackage{mathtools}
\usepackage{bm}
\usepackage{algorithmic}
\usepackage{algorithm}
\usepackage{graphicx}
\usepackage{textcomp}
\usepackage{xcolor}
\usepackage{booktabs}
\usepackage{multirow}
\usepackage{array}
\usepackage{tabularx}
\usepackage{subcaption}
\usepackage{url}
\usepackage[hidelinks,
            pdftitle={Bilateral Teleoperation with Compliant 6-DOF Pose-and-Force Sensing},
            pdfauthor={Yue Feng, Weicheng Huang, I-Ming Chen},
            pdfkeywords={bilateral teleoperation, haptic interaction, impedance control, admittance control, compliant force sensing, contact-rich manipulation}]{hyperref}

\graphicspath{{figures/}}
\begin{document}

\title{Bilateral Teleoperation with\\
Compliant 6-DOF Pose-and-Force Sensing
\thanks{Preprint. \copyright~2026 The Authors.}}

\author{\IEEEauthorblockN{Yue Feng}
\IEEEauthorblockA{\textit{Robotics Research Centre} \\
\textit{Nanyang Technological University}\\
Singapore \\
yue011@e.ntu.edu.sg}
\and
\IEEEauthorblockN{Weicheng Huang}
\IEEEauthorblockA{\textit{WinGs Robotic}\\
USA \\
info@wingsrobotics.com}
\and
\IEEEauthorblockN{I-Ming Chen}
\IEEEauthorblockA{\textit{Robotics Research Centre} \\
\textit{Nanyang Technological University}\\
Singapore \\
michen@ntu.edu.sg}
}

\maketitle

\begin{abstract}
Existing bilateral teleoperation platforms still rely on costly
rigid six-axis force/torque sensors, tightly coupled leader--follower
hardware, and kilohertz control loops. We present a Cartesian
bilateral framework built on the hardware-agnostic WinGs Operating
Studio (WOS) middleware, in which a low-cost compliant 6-DOF
pose-and-force sensing end-effector, Delta6, is mounted on both
sides so that each manipulator behaves as an end-effector
6-DOF series elastic actuator (SEA). The leader runs a damping-only
admittance loop with a 6-D biquad notch filter; the follower
realizes a stiffness--damping impedance through a position-based
outer loop with a PID wrench-to-pose mapping. Three time scales
(hardware I/O, mid-rate impedance/admittance, low-rate teleoperation
messages) are explicitly decoupled, enabling the same application
to drive heterogeneous arms. On a Lite6/FR3 testbed at 150~Hz,
the system tracks stably under delays up to $120{\pm}40$~ms and
1\% packet loss, matches the prescribed virtual stiffness in
contact, and shows a favorable cumulative energy signature in
passivity-style tests.
\end{abstract}


\section{Introduction}
Contact-rich manipulation tasks such as insertion, deburring,
polishing, wiping, and sealing require sustained or transient
interaction with the environment. Achieving stable perception and
control under contact has emerged as a fundamental challenge as
robots move from constructed laboratories toward semi-structured
environments. When a system relies only on vision and proprioception,
occlusion, specular reflection, loss of texture, and sensor latency
create perceptual lag at the instant of contact, while proprioceptive
signals alone do not directly reveal environmental reaction
forces~\cite{suomalainen2022contactsurvey}.

A pragmatic engineering path is to couple bilateral teleoperation
with a closed data loop: even before fully autonomous policies are
deployed, teleoperation delivers immediate operational value while
simultaneously accumulating high-quality visuo-haptic demonstrations
for learning-from-demonstration (LfD)~\cite{ravichandar2020recent}.
A hidden advantage is that the operator's perceptual channel can be
time-synchronized with the observation stream accessible to the
learning system, ensuring that the recorded observations are
sufficient to reproduce interactive policies.

Three barriers, however, continue to limit the scalability of haptic
teleoperation. First, hardware heterogeneity and fragmented
interfaces cause leader--follower devices to remain tightly coupled.
Remote operators are typically required to use equipment identical
to that on site, which prevents broad participation and impedes
large-scale dataset accumulation. Second, mainstream haptic chains
rely on costly rigid strain-gauge six-axis force/torque (F/T)
sensors and demand kilohertz-rate closed loops to avoid rigid
impacts~\cite{hulin2013passivity,Panzirsch2022ScienceRobotics},
imposing stringent requirements on actuators, sensors, and
computing. Third, even when sensing is in place, latency, jitter,
and packet loss over networked links readily destabilize a tightly
coupled bilateral controller~\cite{kamtam2024teleoplatency,farajiparvar2020delay,hespanha2007ncs}.

This paper presents a bilateral teleoperation framework that
explicitly attacks these three barriers, with the following
contributions:
\begin{enumerate}
\item A \emph{hardware-agnostic, Cartesian-space bilateral
architecture} built on the WinGs Operating Studio
(WOS)~\cite{feng2024wos} middleware, in which a low-cost compliant
6-DOF end-effector force sensor, Delta6~\cite{feng2026delta6}, is
mounted on both leader and
follower so that each arm behaves as a 6-DOF end-effector series
elastic actuator (SEA)~\cite{Pratt1995SEA};
\item A \emph{damping-only Cartesian admittance loop with a 6-D
biquad notch filter}~\cite{BristowJohnson2002} on the leader side
that attenuates gain peaking induced by manipulator bandwidth and
structural resonance, and a \emph{stiffness--damping Cartesian
impedance loop} on the follower side realized through a
position-based outer loop with a PID wrench-to-pose mapping;
\item An explicit \emph{three-rate decoupling} (hardware I/O,
medium-rate impedance/admittance, low-rate teleoperation messages)
that, in combination with WOS's unified Cartesian execution
interface, allows the same application to drive heterogeneous
leader and follower platforms across an emulated wide-area link;
\item \emph{Quantitative bilateral validation} including
unilateral Bode characterization under four network conditions,
collision stability and steady-state compliance verification, and
a passivity-style energy analysis on a Lite6/FR3 testbed.
\end{enumerate}

\section{Related Work}
\noindent\textbf{Bilateral teleoperation platforms.}
Recent haptic leader--follower platforms commonly adopt high-end
Force Dimension masters (Sigma.7, Omega.6) and rely either on
rigid wrist 6-axis F/T sensors (e.g., ATI, OptoForce, AIDIN) or on
joint-torque estimation. Representative examples include RH20T,
REASSM, RoboMAN, TCFIC, and PDLOR, which deliver precise haptic
rendering but at high procurement and integration
cost~\cite{fang2023rh20t,reassemble2025,wang2025roboman,kouhkiloui2024telecoopfic,chen2023machines}.
One-to-one twin-arm systems such as TwinPanda achieve high
consistency through identical hardware but remain costly and infer
the end-effector wrench from joint torques, which is sensitive to
motion-induced noise~\cite{singh2020twinpanda}. Low-cost stacks such
as FACTR and Echo demonstrate that bilateral teleoperation is
feasible with modest hardware, but the degrees and accuracy of
force feedback are limited~\cite{liu2025factr,bazhenov2025echo}.
Across these systems, hardware coupling between leader and follower
remains strong: changing one side typically requires re-implementing
integration interfaces and revalidating the entire stack.

\noindent\textbf{Compliant 6-DOF force sensing.}
Rigid metal strain-gauge F/T sensors set the accuracy and bandwidth
standard but are expensive and impose safety
risks during dynamic contact~\cite{Panzirsch2022ScienceRobotics}.
Research efforts on low-cost optical, photo-interrupter, and
fiducial-based 6-DOF sensors have either delivered modest accuracy,
or have required careful calibration and
alignment~\cite{Hendrich2020Access,Palli2014Optoelectronic,AlMai2018Optical6axis,Ouyang2020Fiducial}.
The Delta6 design used in this work~\cite{feng2026delta6} is a
compliant, spring-driven, magnetic-encoder-based 6-DOF F/T
end-effector that achieves usable accuracy at a fraction of the
cost of commercial transducers, with mechanical compliance that
enables stable contact at sub-kilohertz control rates.

\noindent\textbf{Networked impedance/admittance teleoperation.}
Energetic stability under communication delay has been studied
extensively, with passivity-based and wave-variable schemes
providing the dominant theoretical
framework~\cite{hulin2013passivity}. In practice, however,
empirical performance under realistic networked conditions remains
limited by the available motion-component bandwidth and by the
quality of the middleware time and rate
contracts~\cite{kamtam2024teleoplatency,bray2024latencycomposition}.
The architecture proposed here complements that line of work by
exposing the teleoperation pipeline as a multi-rate cascade
implemented on a portable, hardware-agnostic middleware.

\section{System Architecture}
\label{sec:arch}
\subsection{Overview}
\begin{figure}[t]
  \centering
  \includegraphics[width=\linewidth]{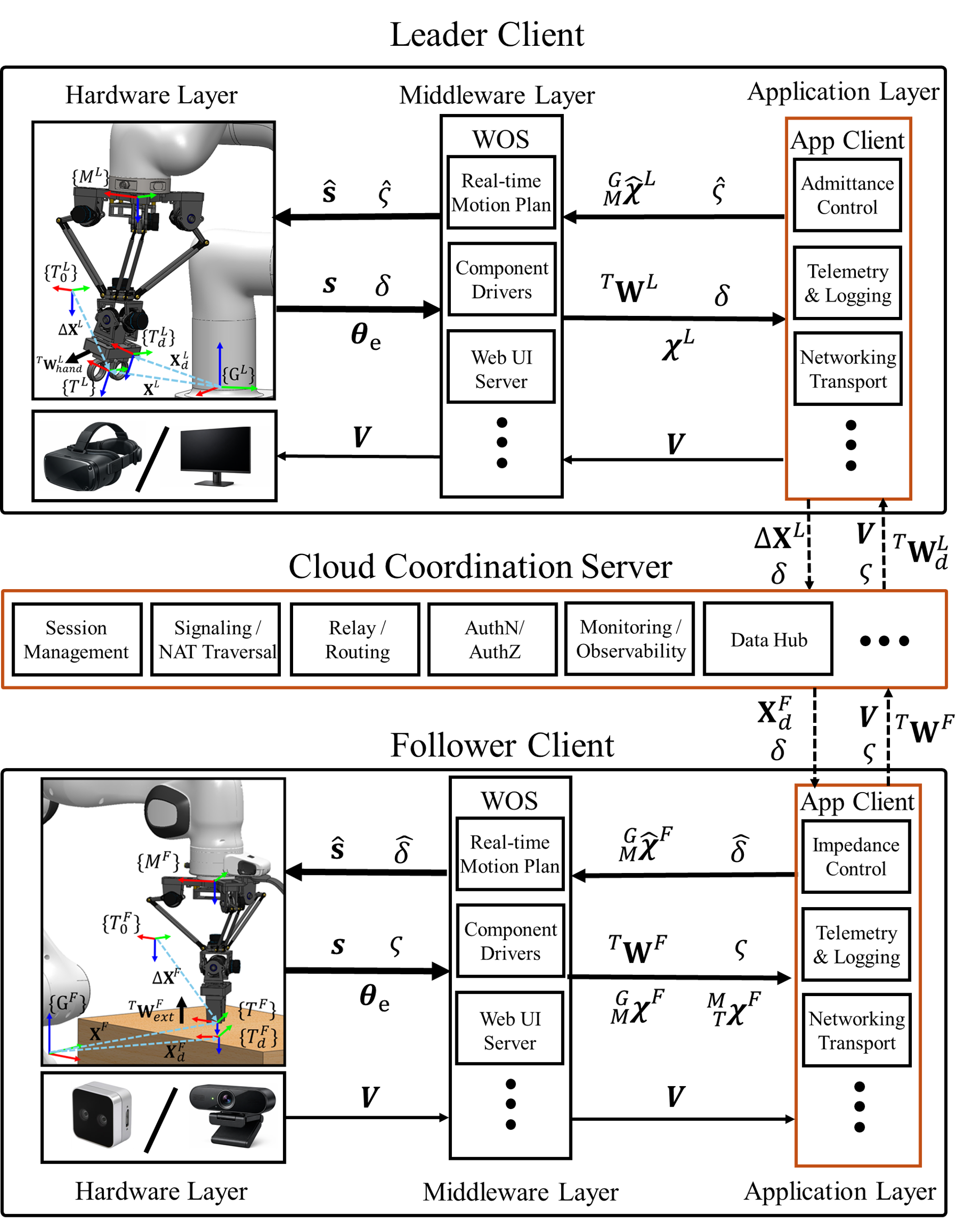}
  \caption{Data pipeline of the bilateral leader--follower
  teleoperation framework. WOS exposes a unified Cartesian
  command/state interface for both heterogeneous arms; Delta6
  provides 6-DOF wrench and pose at the end-effector on both
  sides. Teleoperation messages traverse a server that performs
  session, NAT, AuthN/AuthZ, and storage functions.}
  \label{fig:teleop_pipeline}
\end{figure}

The proposed bilateral system, depicted in
Fig.~\ref{fig:teleop_pipeline}, consists of three layers. The
\emph{device layer} comprises 6-DOF manipulators, the Delta6
6-DOF compliant end-effector sensors, grippers, and visual
sensors. The \emph{middleware layer} (WOS) abstracts heterogeneous
devices into a unified network API and embeds a real-time
Cartesian motion-planning service that exposes a single
\texttt{rt-move-cartesian} call accepting offline trajectories,
interruptible waypoints, and real-time teleoperation setpoints
under bounded per-cycle computation. The \emph{application layer}
implements the bilateral control loops and the visual stream
relay; it interacts only with the middleware API and is therefore
portable across heterogeneous leader/follower platforms.

\subsection{Minimal Viable Component Sets}
To make hardware substitution explicit, the framework assumes the
following minimal viable component sets:
\begin{itemize}
\item \textbf{Leader.} A 6-DOF motion component with Cartesian
state/command, a 6-DOF wrench/pose sensing component (e.g., Delta6
or any rigid 6-axis F/T transducer with constant flange--TCP
transform), a gripper controller that accepts a force reference
$\hat{\varsigma}$ and returns the gripper position $\delta$, and a
display rendering the incoming video stream.
\item \textbf{Follower.} A 6-DOF motion component, a 6-DOF
wrench/pose sensor, a gripper that accepts a position reference
$\hat{\delta}$ and returns the measured interaction force
$\varsigma$, and a camera providing RGB(D) frames with a known
extrinsic to the flange.
\end{itemize}
For compliant sensors such as Delta6, the relative flange--TCP
pose is non-constant and is estimated online by WOS; for rigid
sensors, the transform is constant.

\subsection{Three-Rate Decoupling}
Three time scales are intentionally separated. Hardware I/O inside
WOS runs at the robot control frequency $f_c$ (typically several
hundred Hz). The Cartesian admittance/impedance loops on the
clients run at $f^{L}_{\mathrm{admt}}$ and $f^{F}_{\mathrm{impd}}$
(typically $\sim$150~Hz). Teleoperation messages traveling along
leader-client $\rightarrow$ server $\rightarrow$ follower-client
run at $f_{\mathrm{tele}}$ (typically $\sim$50~Hz) and tolerate
latency, jitter, and packet loss. Robustness is supported by
time-stamped messages, latest-sample hold between updates, and
watchdog-based safe fallback when updates are missing beyond a
timeout.

\section{Bilateral Cartesian Control}
\label{sec:control}
\subsection{Notation}
\begin{figure}[t]
  \centering
  \includegraphics[width=0.95\linewidth]{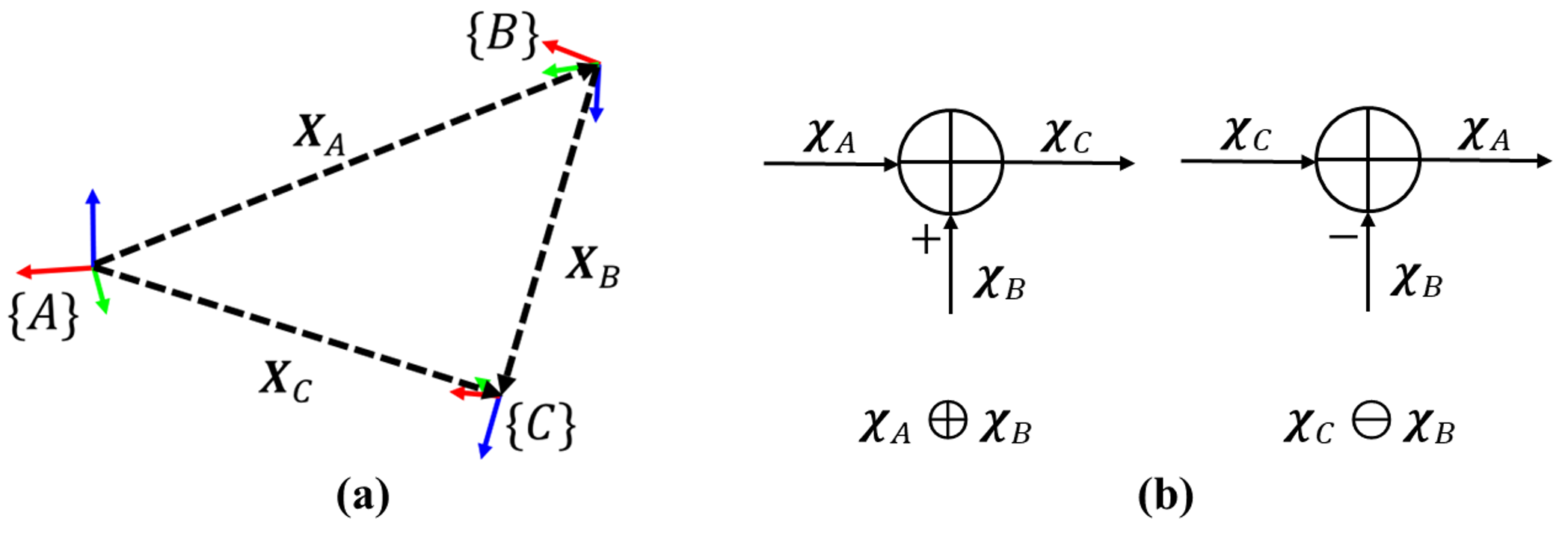}
  \caption{Kinematic composition operators $\oplus$ and $\ominus$:
  (a) geometric interpretation in $\mathrm{SE}(3)$;
  (b) shorthand used in the control diagrams.}
  \label{fig:pose_comp}
\end{figure}

A Cartesian pose is parameterized as
$\mathbf{X}=[x,y,z,\phi,\vartheta,\psi]^{\mathsf T}$ with XYZ Euler
angles, with the homogeneous transform
$\mathbf{T}\triangleq\mathcal{T}(\mathbf{X})\in\mathrm{SE}(3)$ and
inverse $\mathbf{X}=\operatorname{vec}_{XYZ}(\mathbf{T})$. We
group a frame state as
$\boldsymbol{\chi}\triangleq(\mathbf{X},\dot{\mathbf{X}},
\ddot{\mathbf{X}})$ and define kinematic composition operators
$\oplus,\ominus$ acting on $\boldsymbol{\chi}$ that compose poses
together with consistent twist/acceleration transport
(Fig.~\ref{fig:pose_comp}). Superscripts $L,F$ denote leader and
follower; a left super/subscript pair, e.g.,
$\prescript{G}{M}{\mathbf{T}}$, indicates a transform from frame
$M$ to frame $G$. We write $\prescript{T}{}{\mathbf{W}}$ for a
wrench expressed in the TCP frame $\{T\}$.

\subsection{Leader-Side Admittance Loop}
\begin{figure}[t]
  \centering
  \includegraphics[width=0.85\linewidth]{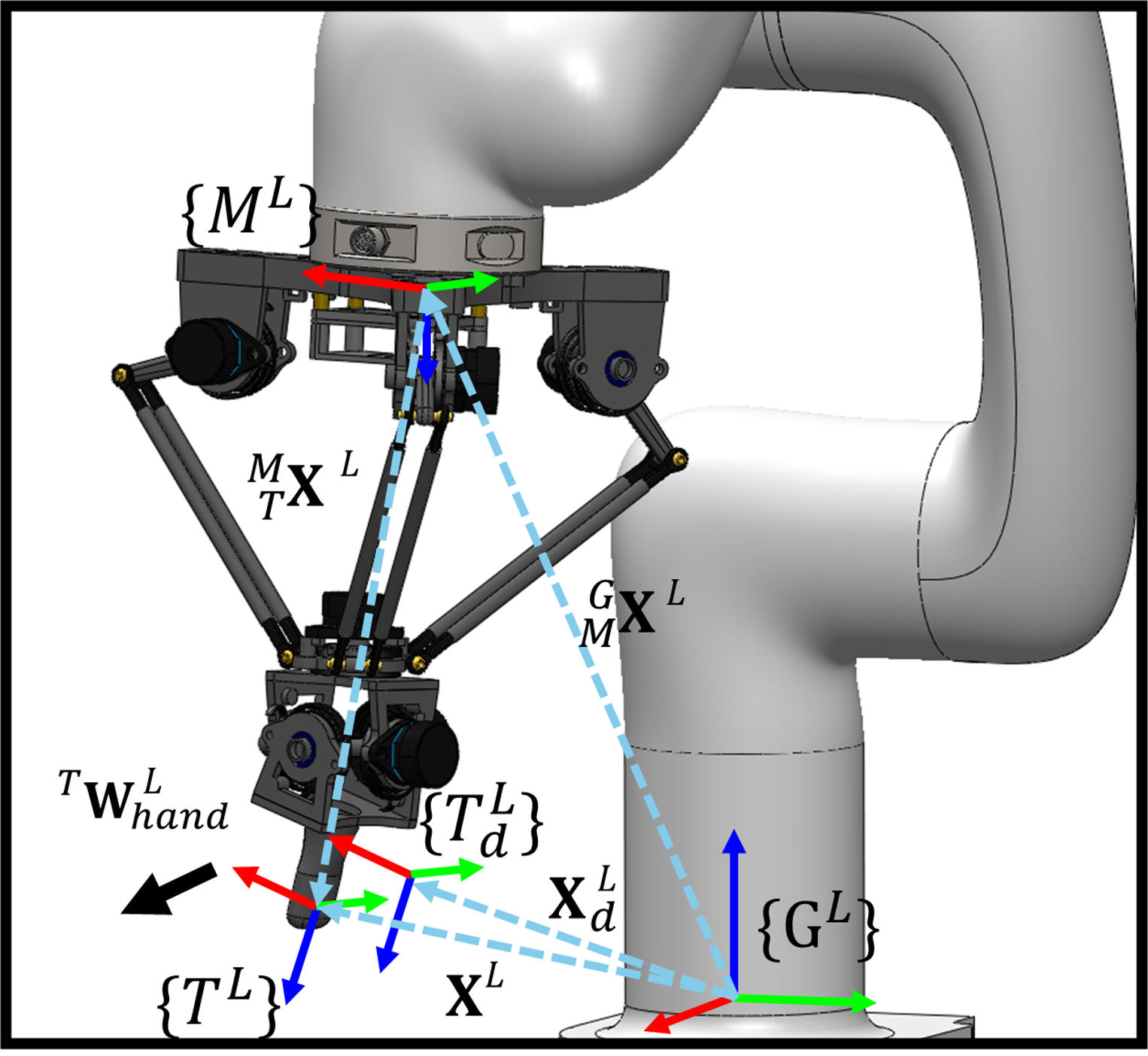}
  \caption{Leader-side coordinate frames and Delta6 mounting on
  the 6-DOF manipulator.}
  \label{fig:leader_frames}
\end{figure}
A Delta6 is mounted at the flange of the leader-side 6-DOF
manipulator (Fig.~\ref{fig:leader_frames}). Let $\{G^L\}$,
$\{M^L\}$, $\{T^L\}$ denote the leader global, flange and TCP
frames. The human operator applies a wrench
$\prescript{T}{}{\mathbf{W}}^{L}_{\mathrm{hand}}$ to $\{T^L\}$,
which is measured by Delta6 as
$\prescript{T}{}{\mathbf{W}}^{L}$. With Cartesian inertia,
damping, and stiffness $\mathbf{M}^{L}$, $\mathbf{B}^{L}$,
$\mathbf{K}^{L}$, the standard second-order admittance reads
\begin{equation}
\mathbf{M}^{L}\ddot{\mathbf{X}}^{L}_{d}
+\mathbf{B}^{L}\dot{\mathbf{X}}^{L}_{d}
+\mathbf{K}^{L}\!\bigl(\mathbf{X}^{L}_{d}-\mathbf{X}^{L}_{0}\bigr)
=\prescript{T}{}{\mathbf{W}}^{L}_{\mathrm{hand}}.
\label{eq:adm_standard}
\end{equation}
For simplicity we adopt a damping-only formulation,
\begin{equation}
\mathbf{B}^{L}\dot{\mathbf{X}}^{L}_{d}
=\prescript{T}{}{\mathbf{W}}^{L}_{\mathrm{hand}}
-\prescript{T}{}{\breve{\mathbf{W}}}^{L}_{d},
\label{eq:adm_damping_only}
\end{equation}
where $\prescript{T}{}{\breve{\mathbf{W}}}^{L}_{d}$ is the
filtered follower-side reference wrench used for haptic feedback.
The wrench error is converted into a commanded twist
${}^{T}\dot{\mathbf{X}}^{L}_{d}=(\mathbf{B}^{L})^{-1}\,
{}^{T}\mathbf{e}_{W}$, integrated with forward Euler in
translation, and integrated through the XYZ Euler-rate Jacobian
$\mathbf{J}_{XYZ}(\boldsymbol{\varphi})$ in orientation. The pose
update is then applied through the kinematic composition operator,
${}^{T}\mathbf{X}^{L}_{d}\leftarrow {}^{T}\mathbf{X}^{L}_{d}\oplus
\Delta {}^{T}\mathbf{X}^{L}_{d}$.

\begin{figure}[t]
  \centering
  \includegraphics[width=\linewidth]{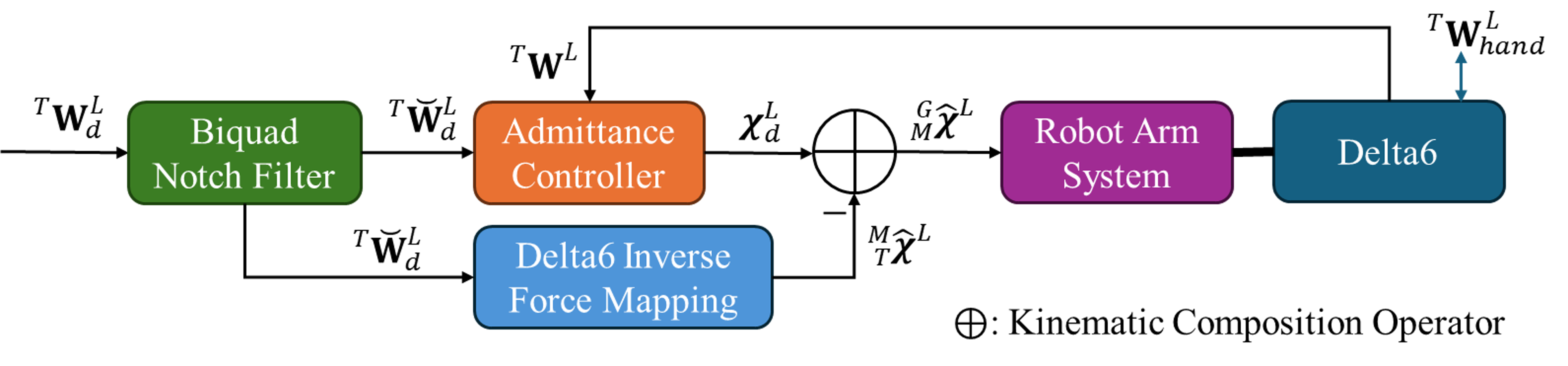}
  \caption{Block diagram of the Delta6-based Cartesian admittance
  control loop on the leader side.}
  \label{fig:leader_block}
\end{figure}

\noindent\textbf{6-D biquad notch filter (BNF).}
The reference wrench $\prescript{T}{}{\mathbf{W}}^{L}_{d}$
streamed back from the follower is processed through a 6-D
RBJ-form biquad notch filter~\cite{BristowJohnson2002} applied
independently to each wrench component. The filter attenuates
gain peaking arising from the finite bandwidth and phase lag of
the manipulator inner loop, structural flexibility and resonant
modes, and aggregation of closed-loop gain near particular
frequencies. For axis $i$ with notch center $f_{0,i}$ and
sharpness $\varkappa_i$, normalized RBJ coefficients are computed
from $\omega_0=2\pi f_{0,i}/f^{L}_{\mathrm{admt}}$ and
$\eta=\sin\omega_0/(2\varkappa_i)$; a per-axis depth coefficient
$\lambda_i\in[0,1]$ blends the raw biquad output back toward the
unfiltered signal,
\begin{equation}
\boldsymbol{\upsilon}=(\mathbf{1}-\boldsymbol{\lambda})\odot
\boldsymbol{\varpi}+\boldsymbol{\lambda}\odot
\boldsymbol{\upsilon}^{\circ},
\end{equation}
so that $\lambda_i=0$ bypasses axis $i$ and $\lambda_i=1$ applies
the full notch response.

\noindent\textbf{Delta6 inverse force mapping.}
Because Delta6 acts as a 6-DOF SEA, the filtered desired wrench
$\prescript{T}{}{\breve{\mathbf{W}}}^{L}_{d}$ is converted into a
target end-effector pose by inverting the Delta6 forward-wrench
map $\mathcal{M}(\boldsymbol{\tau})$. Given the residual
$\boldsymbol{\mathfrak r}(\boldsymbol{\tau})\triangleq
\mathcal{M}(\boldsymbol{\tau})-
\prescript{T}{}{\breve{\mathbf{W}}}^{L}_{d}$ a root solver yields
$\boldsymbol{\tau}^\star$; spring stiffness $k_s^L$ converts
torques to angles, and forward kinematics yields the next-tick
target Delta6 state
$\prescript{M}{T}{\hat{\boldsymbol{\chi}}}^{L}$. The
admittance-consistent desired TCP state
$\boldsymbol{\chi}^{L}_{d}$ and
$\prescript{M}{T}{\hat{\boldsymbol{\chi}}}^{L}$ are then composed
via $\ominus$ to form the next-tick manipulator command
$\prescript{G}{M}{\hat{\boldsymbol{\chi}}}^{L}$
(Fig.~\ref{fig:leader_block}).

\subsection{Follower-Side Impedance Loop}
\begin{figure}[t]
  \centering
  \includegraphics[width=\linewidth]{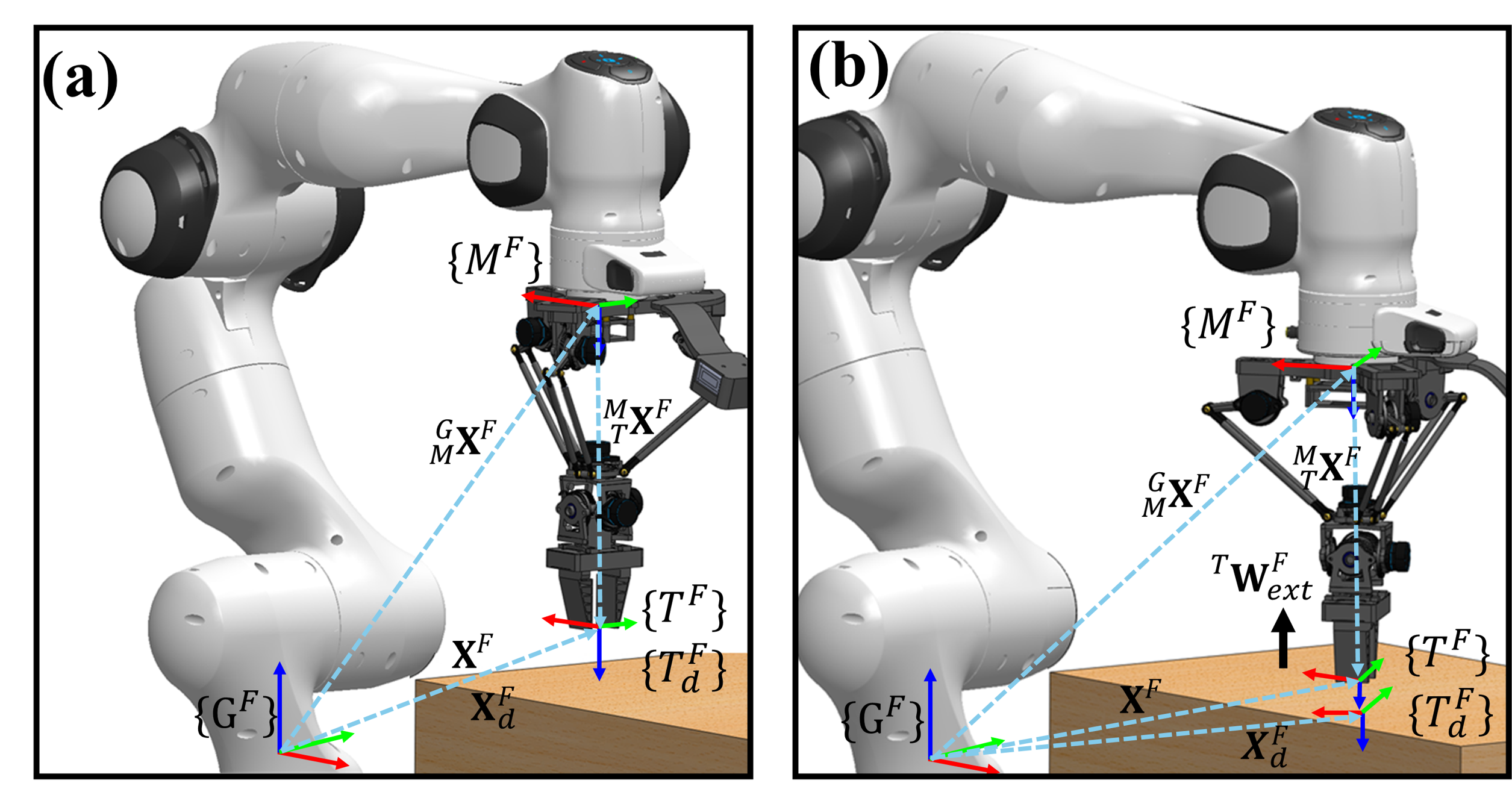}
  \caption{Follower-side coordinate frames and Delta6 mounting on
  the 6-DOF manipulator. (a) free motion; (b) contact interaction
  with an external wrench.}
  \label{fig:follower_frames}
\end{figure}
On the follower side, a Delta6 is mounted on the manipulator
flange (Fig.~\ref{fig:follower_frames}). The desired Cartesian
behavior at the TCP is a virtual mass--spring--damper relating
the pose error to the external wrench
$\prescript{T}{}{\mathbf{W}}^{F}_{\mathrm{ext}}$,
\begin{equation}
\mathbf{M}^{F}\ddot{\mathbf{X}}^{F}
+\mathbf{B}^{F}\dot{\mathbf{X}}^{F}
+\mathbf{K}^{F}\!\bigl(\mathbf{X}^{F}-\mathbf{X}^{F}_{d}\bigr)
=\prescript{T}{}{\mathbf{W}}^{F}_{\mathrm{ext}}.
\label{eq:impd_standard}
\end{equation}
We retain the stiffness--damping terms and neglect inertia,
\begin{equation}
\mathbf{B}^{F}\dot{\mathbf{X}}^{F}
+\mathbf{K}^{F}\!\bigl(\mathbf{X}^{F}-\mathbf{X}^{F}_{d}\bigr)
=\prescript{T}{}{\mathbf{W}}^{F}_{\mathrm{ext}}.
\label{eq:impd_BK}
\end{equation}
In free motion the damping shapes the velocity response; under
contact the stiffness produces the restoring wrench
(Fig.~\ref{fig:follower_frames}).

\begin{figure}[t]
  \centering
  \includegraphics[width=\linewidth]{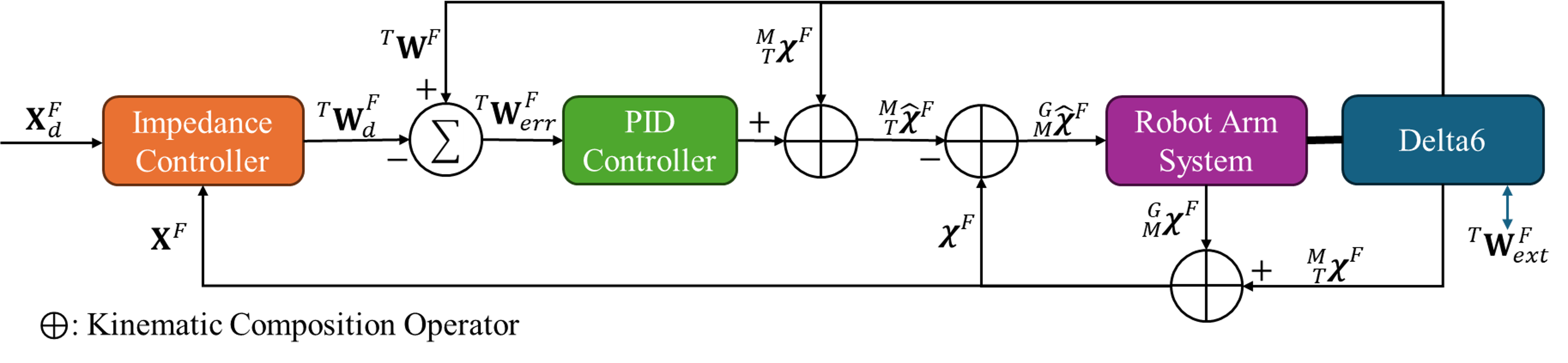}
  \caption{Block diagram of the Delta6-based Cartesian impedance
  control loop on the follower side.}
  \label{fig:follower_block}
\end{figure}

Because native 6-DOF Cartesian wrench control is typically not
exposed by collaborative arms, the impedance is realized through
a position-based outer loop (Fig.~\ref{fig:follower_block}).
In each control cycle the impedance block computes a reference
wrench from the pose error and a filtered velocity error,
\begin{equation}
\prescript{T}{}{\mathbf{W}}^{F}_{d}=\mathbf{K}^{F}\odot
\mathbf{e}_{p}+\mathbf{B}^{F}\odot\breve{\mathbf{e}}_{v},
\end{equation}
where $\mathbf{e}_p$ is the $\mathrm{SE}(3)$ pose deviation
expressed as a 6-vector via $\operatorname{vec}_{XYZ}(\cdot)$ and
$\breve{\mathbf{e}}_v$ is a first-order low-pass filtered finite
difference (time constant $\tau_v$). The wrench error
$\prescript{T}{}{\mathbf{W}}^{F}_{\mathrm{err}}=
\prescript{T}{}{\mathbf{W}}^{F}-
\prescript{T}{}{\mathbf{W}}^{F}_{d}$ then drives a 6-D PID
controller that produces an incremental pose command
$\Delta\prescript{M}{T}{\mathbf{X}}^{F}=\mathbf{k}_p\odot
\prescript{T}{}{\mathbf{W}}^{F}_{\mathrm{err}}+\mathbf{k}_i\odot
\mathbf{I}+\mathbf{k}_d\odot\mathbf{D}$. Composing the increment
with the measured Delta6 state and the flange state via $\oplus$
and $\ominus$ yields the next-tick flange command
\begin{equation}
\prescript{G}{M}{\hat{\boldsymbol{\chi}}}^{F}
=\bigl(\prescript{G}{M}{\boldsymbol{\chi}}^{F}
\oplus\prescript{M}{T}{\boldsymbol{\chi}}^{F}\bigr)
\ominus\prescript{M}{T}{\hat{\boldsymbol{\chi}}}^{F},
\label{eq:follower_cmd}
\end{equation}
which is sent to the manipulator through WOS as the target state
for the next servo update.

\subsection{Teleoperation Coupling}
At the leader client, the relative TCP increment is
\begin{equation}
\Delta\mathbf{X}^{L}=\operatorname{vec}_{XYZ}\!\Bigl(
\mathcal{T}(\mathbf{X}^{L})^{-1}\,
\mathcal{T}(\mathbf{X}^{L}_{0})\Bigr).
\label{eq:dxL}
\end{equation}
Time-stamped setpoints $\Delta\mathbf{X}^{L}$ are transmitted at
$f_{\mathrm{tele}}$ to the server and forwarded to the follower
client, where the desired TCP pose is reconstructed as
\begin{equation}
\mathbf{X}^{F}_{d}=\mathbf{X}^{F}_{0}\oplus\Delta\mathbf{X}^{L}
\label{eq:xFd}
\end{equation}
and fed into the impedance loop. Bilateral haptic feedback is
closed by streaming the follower external wrench
$\prescript{T}{}{\mathbf{W}}^{F}_{\mathrm{ext}}$ back through the
server to serve as the leader reference
$\prescript{T}{}{\mathbf{W}}^{L}_{d}$. The gripper channel follows
complementary semantics: the leader writes $\hat{\varsigma}$ and
reads $\delta$, while the follower writes $\hat{\delta}$ and reads
$\varsigma$. The video stream is acquired by WOS on the follower,
forwarded by the server, and rendered on the leader (HMD or
monitor) as RGB or RGB-D.

\section{Experimental Validation}
\label{sec:exp}
\subsection{Prototype and Network Emulation}
\begin{figure}[t]
  \centering
  \includegraphics[width=\linewidth]{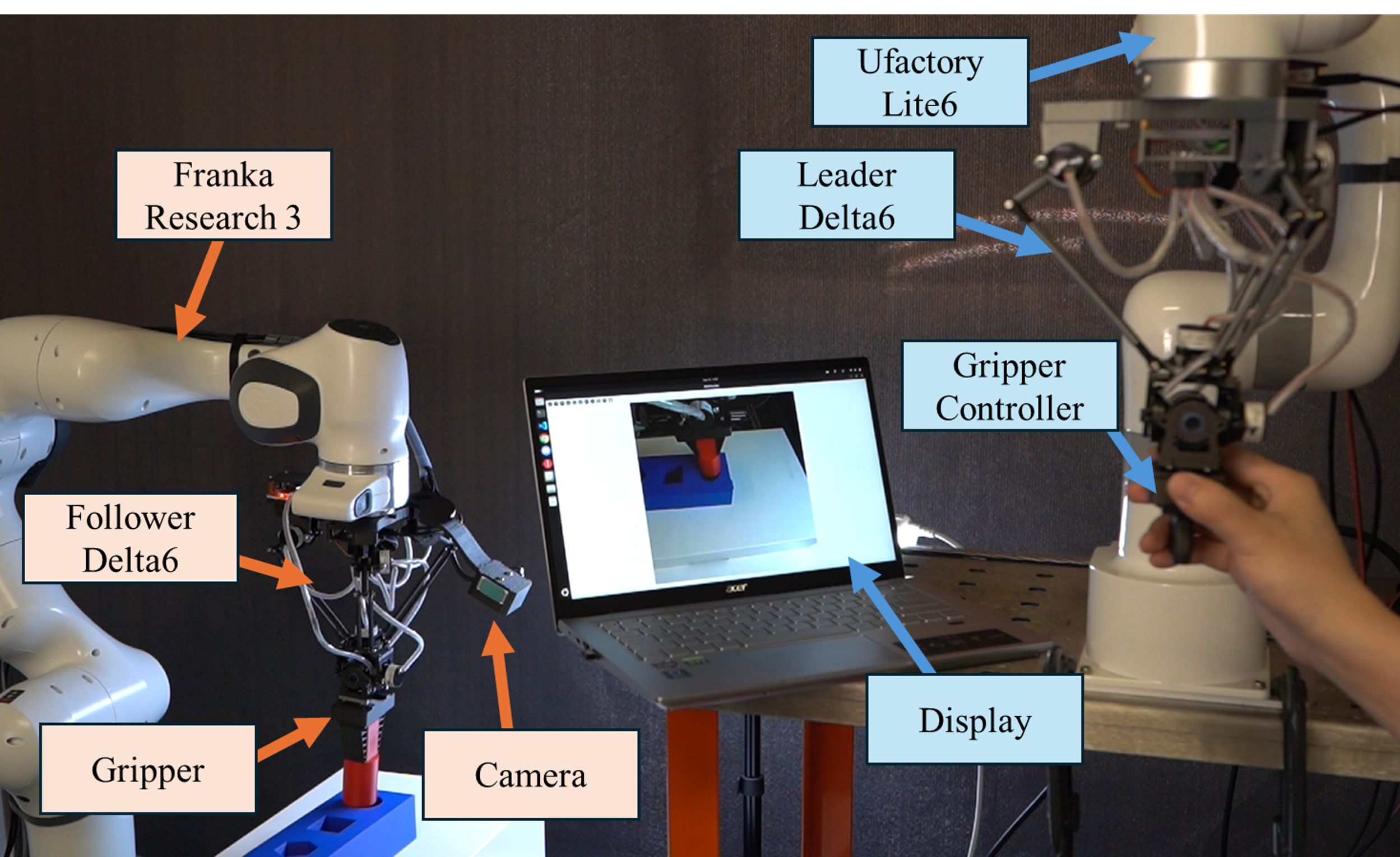}
  \caption{Local bilateral testbed: follower side (left, yellow)
  and leader side (right, blue). Both sides employ a Delta6 module
  as a 6-DOF wrench sensor. The leader uses a UFactory Lite6, and
  a RealSense D405 RGB-D camera on the follower provides RGB
  video to the operator.}
  \label{fig:testbed}
\end{figure}

\begin{table}[t]
\centering
\caption{Emulated network conditions.}
\label{tab:net}
\setlength{\tabcolsep}{4pt}
\renewcommand{\arraystretch}{1.10}
\begin{tabular}{lccc}
\toprule
Condition & Mean Delay (ms) & Std.\ (ms) & Loss (\%) \\
\midrule
Local & 0   & 0  & 0.0 \\
Good  & 40  & 10 & 0.0 \\
Fair  & 80  & 20 & 0.5 \\
Poor  & 120 & 40 & 1.0 \\
\bottomrule
\end{tabular}
\end{table}

\begin{table}[t]
\centering
\caption{Controller and timing parameters.}
\label{tab:params}
\setlength{\tabcolsep}{4pt}
\renewcommand{\arraystretch}{1.10}
\begin{tabular}{p{0.28\linewidth} p{0.62\linewidth}}
\toprule
Group / Param. & Value \\
\midrule
$f_c^L, f_c^F$ & 250 Hz \\
$f^{L}_{\mathrm{admt}}, f^{F}_{\mathrm{impd}}$ & 150 Hz \\
$f_{\mathrm{tele}}$ & 50 Hz \\
$\mathbf{B}^{L}$ & $[60,60,100,0.3,0.3,0.2]$ \\
$\mathbf{f}_0$ / $\boldsymbol{\varkappa}$ / $\boldsymbol{\lambda}$ &
  $1$~Hz / $1.2$ / $0.27$ (all axes) \\
$\mathbf{B}^{F}$ & $[2,2,3,0.2,0.2,0.1]$ \\
$\mathbf{K}^{F}$ & $[300,300,500,2.4,2.4,1.2]$ \\
$\tau_v$ & 0.03 s \\
$\mathbf{k}_p$ & $[0.1,0.1,0.25,20,20,40]$ \\
$\mathbf{k}_i$ & $[0.01,0.01,0.02,5,5,10]$ \\
Delta6 $k_s$ & 0.64 N$\cdot$m/rad \\
\bottomrule
\end{tabular}
\end{table}

A minimal local instantiation of Fig.~\ref{fig:teleop_pipeline} is
constructed (Fig.~\ref{fig:testbed}). Both leader and follower
mount a Delta6; the leader uses a UFactory Lite6 with a custom
servo-driven gripper controller, and the follower carries a
RealSense D405 in RGB mode. All hardware connects to a single
Ubuntu workstation through WOS, exposed to the application layer
via WebSocket. The bilateral server is colocated and emulates
network effects (delay, jitter, loss) to approximate long-distance
operation. Four conditions are defined in Table~\ref{tab:net}.
Controller parameters used throughout the evaluation are
summarized in Table~\ref{tab:params}.

\subsection{Leader-Side Unilateral Evaluation}
\begin{figure}[t]
  \centering
  \includegraphics[width=0.45\linewidth]{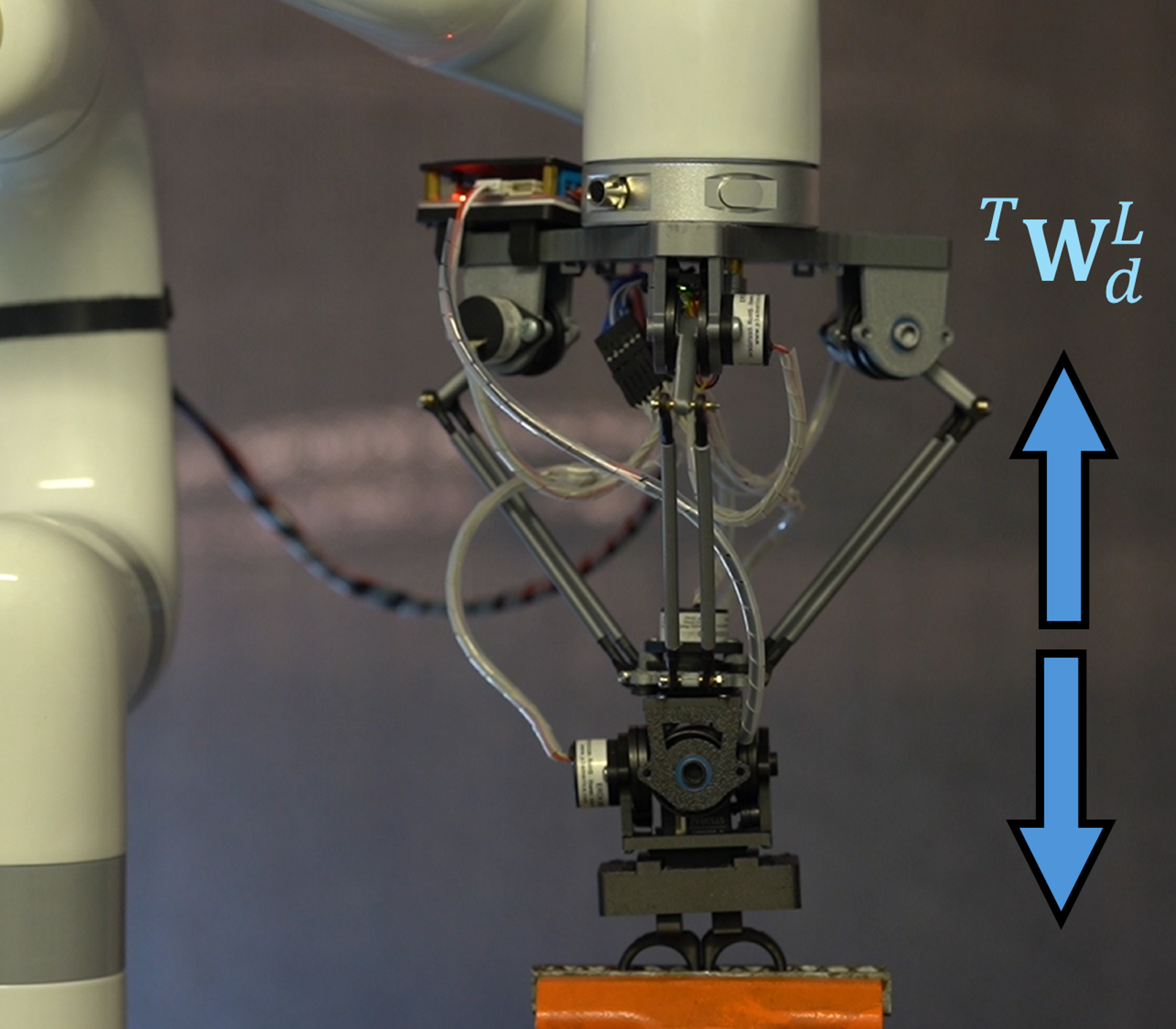}
  \caption{Setup for the leader-side dynamic response: the Delta6
  end-effector is clamped at the neutral pose and excited by a
  sinusoidal tool-frame $z$-axis input of amplitude 1~N at swept
  frequencies.}
  \label{fig:leader_setup}
\end{figure}

To assess wrench tracking, the leader Delta6 is mechanically
clamped at its neutral pose (Fig.~\ref{fig:leader_setup}) and
excited by a reciprocating sinusoidal force of amplitude $1\,$N
along the tool-frame $z$ axis, while the excitation frequency is
swept. Two studies are performed: (i) effect of admittance
damping with the notch filter disabled and (ii) effect of network
condition with the notch filter enabled.

\begin{figure}[t]
  \centering
  \includegraphics[width=\linewidth]{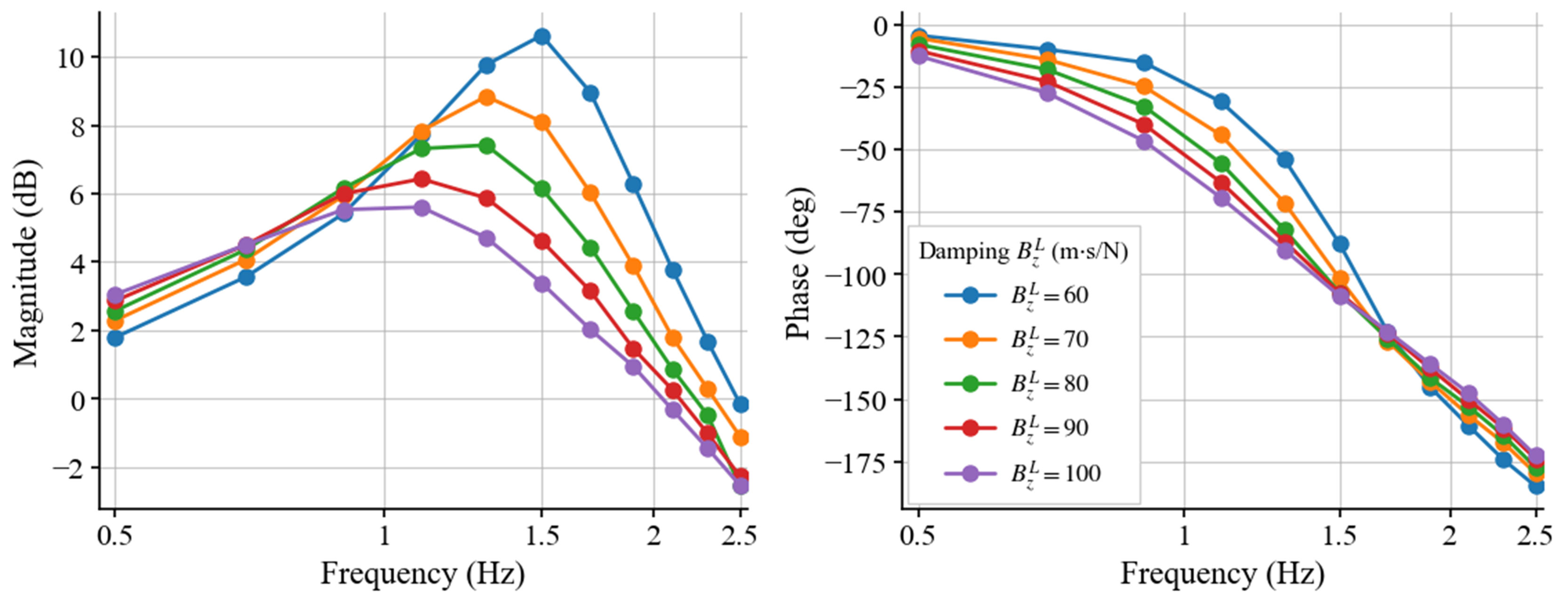}
  \caption{Leader-side admittance Bode plots along $z$ under
  different damping settings without notch filtering
  ($\lambda=0$).}
  \label{fig:leader_bode_no_notch}
\end{figure}

\noindent\textbf{Damping sweep (notch off).}
Fig.~\ref{fig:leader_bode_no_notch} shows the end-to-end
$z$-axis Bode plots obtained by comparing the leader-side measured
wrench $\prescript{T}{}{\mathbf{W}}^{L}$ to the follower-side
injected reference. With low damping $B^L_z=60$, a pronounced
gain peak of $\sim$10~dB appears near 1.5~Hz, and noticeable
oscillations are observed under abrupt stop-and-hold motions,
indicating insufficient damping margin. As $B^L_z$ increases the
gain peak is mitigated and the resonance shifts to lower
frequencies, at the cost of larger phase lag below 1.5~Hz. To
prioritize stability over interaction lightness, $B^L_z=100$ is
adopted in the remaining experiments.

\begin{figure}[t]
  \centering
  \includegraphics[width=\linewidth]{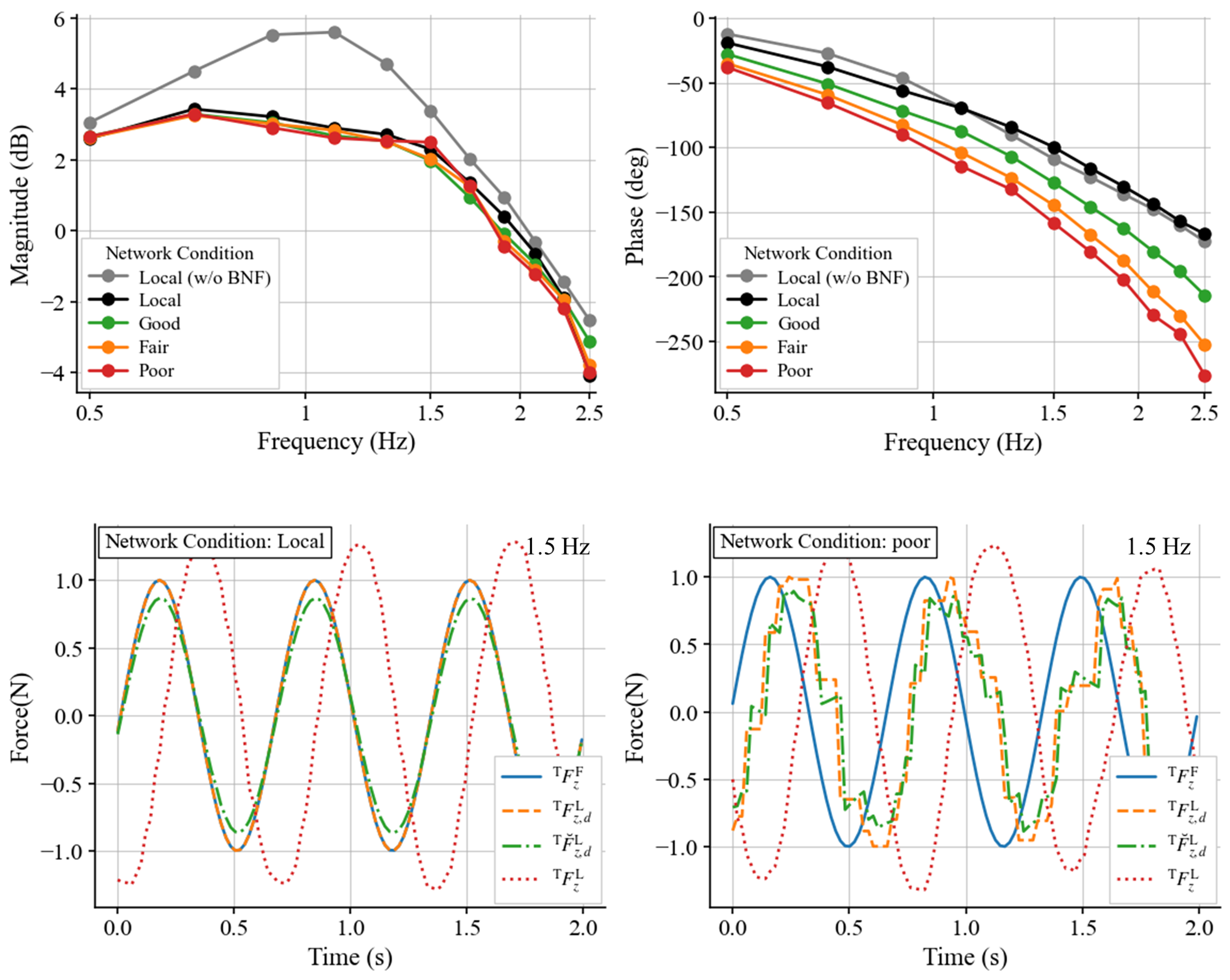}
  \caption{Leader-side admittance frequency and time responses
  with the biquad notch filter enabled ($B^L_z=100$).}
  \label{fig:leader_bode_time}
\end{figure}

\noindent\textbf{Network sweep (notch on).}
Fig.~\ref{fig:leader_bode_time} (top) shows that enabling the
6-D biquad notch filter reduces the dominant gain peak from
$\sim$6~dB to about 3--4~dB, with no noticeable additional phase
lag in the band of interest. Across the four network conditions
the magnitude response is largely unchanged while the phase lag
grows as the channel degrades. Under the \emph{Good} condition,
1.5~Hz excitation exhibits no pronounced gain amplification
($\sim$2~dB) and a phase lag of $\sim$$-125^{\circ}$, equivalent
to an end-to-end delay of $\sim$200~ms, which is adequate for
contact-rich tasks whose completion speed is not critical (e.g.,
polishing, compliant assembly). The time-domain traces in
Fig.~\ref{fig:leader_bode_time} (bottom) at 1.5~Hz confirm that
under the \emph{Poor} condition the reference is visibly distorted
and delayed yet the closed-loop response remains stable, thanks to
the low-pass behavior of admittance damping and the notch filter.

\subsection{Follower-Side Unilateral Evaluation}
\begin{figure}[t]
  \centering
  \includegraphics[width=0.45\linewidth]{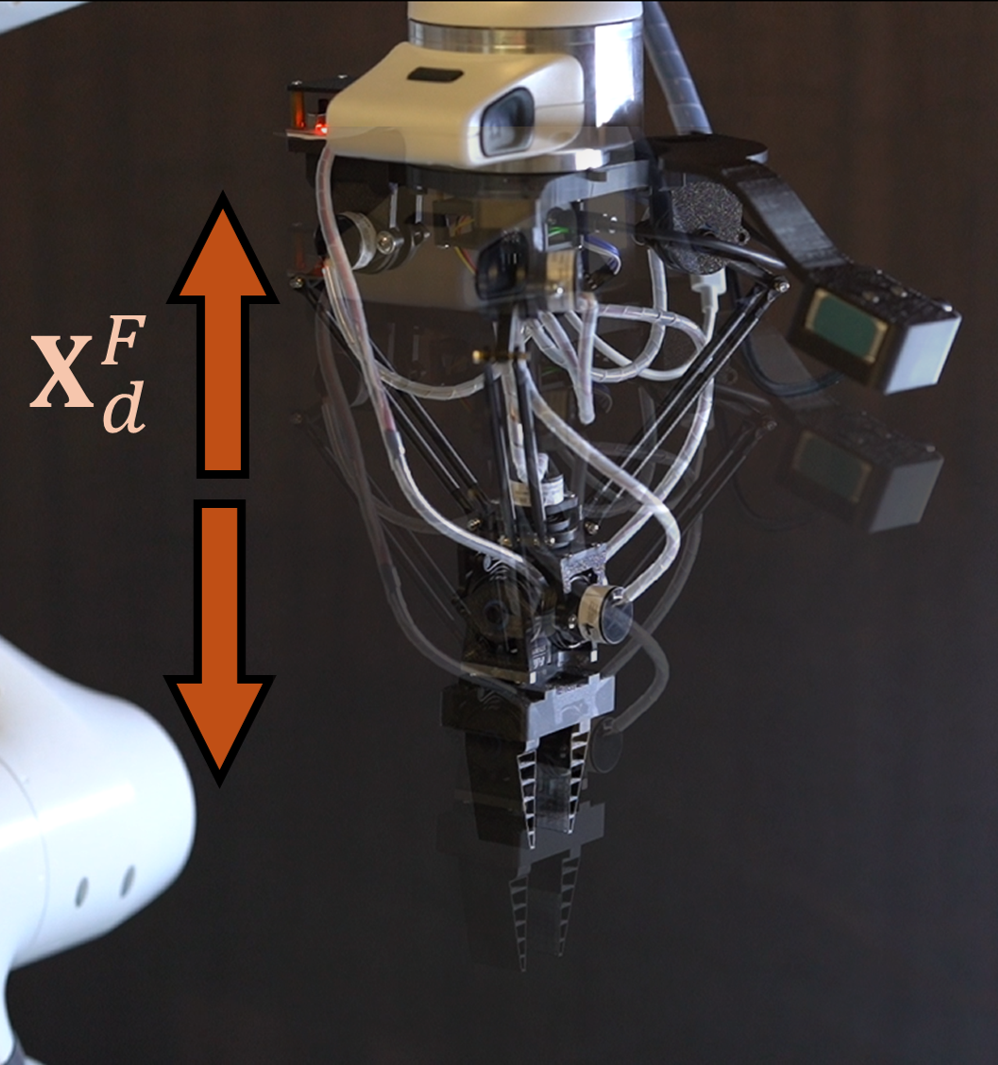}
  \caption{Setup for the follower-side free-space dynamic
  response.}
  \label{fig:follower_setup}
\end{figure}

\begin{figure}[t]
  \centering
  \includegraphics[width=\linewidth]{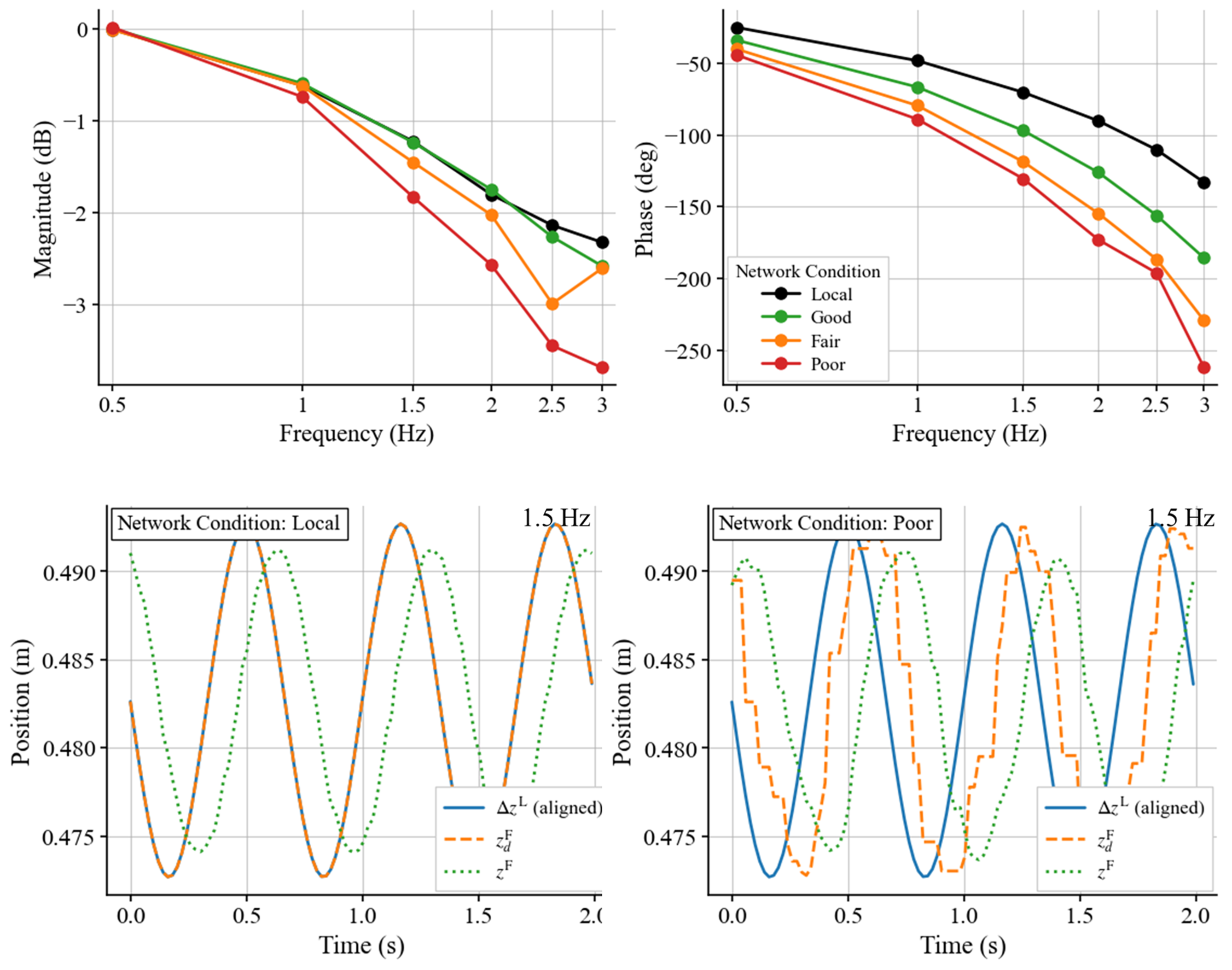}
  \caption{Follower-side impedance frequency and time
  responses across the four network conditions.}
  \label{fig:follower_bode}
\end{figure}

The follower TCP is held in free space near the workspace center
(Fig.~\ref{fig:follower_setup}) and tracked through a sinusoidal
$z$-axis motion command of amplitude 10~mm at swept frequencies.
The Bode plots in Fig.~\ref{fig:follower_bode} (top) are obtained
by comparing the reference relayed from the leader to the measured
follower TCP position. As the network degrades, the $-3\,$dB
bandwidth contracts from above $3\,$Hz under \emph{Local} to
$\sim$$2.2\,$Hz under \emph{Poor}; under the \emph{Good} condition
the combined controller and network delay at 1.5~Hz is $\sim$185~ms.
Even under cloud-emulated conditions the system retains a margin
for everyday manipulation. The time-domain plots in
Fig.~\ref{fig:follower_bode} (bottom) at 1.5~Hz show that the
\emph{Poor} reference signal is significantly distorted by
stochastic delay and packet loss, yet the output motion remains
smooth, owing to the damping of the impedance layer and the
real-time interpolation provided by WOS.

\begin{figure}[t]
  \centering
  \includegraphics[width=\linewidth]{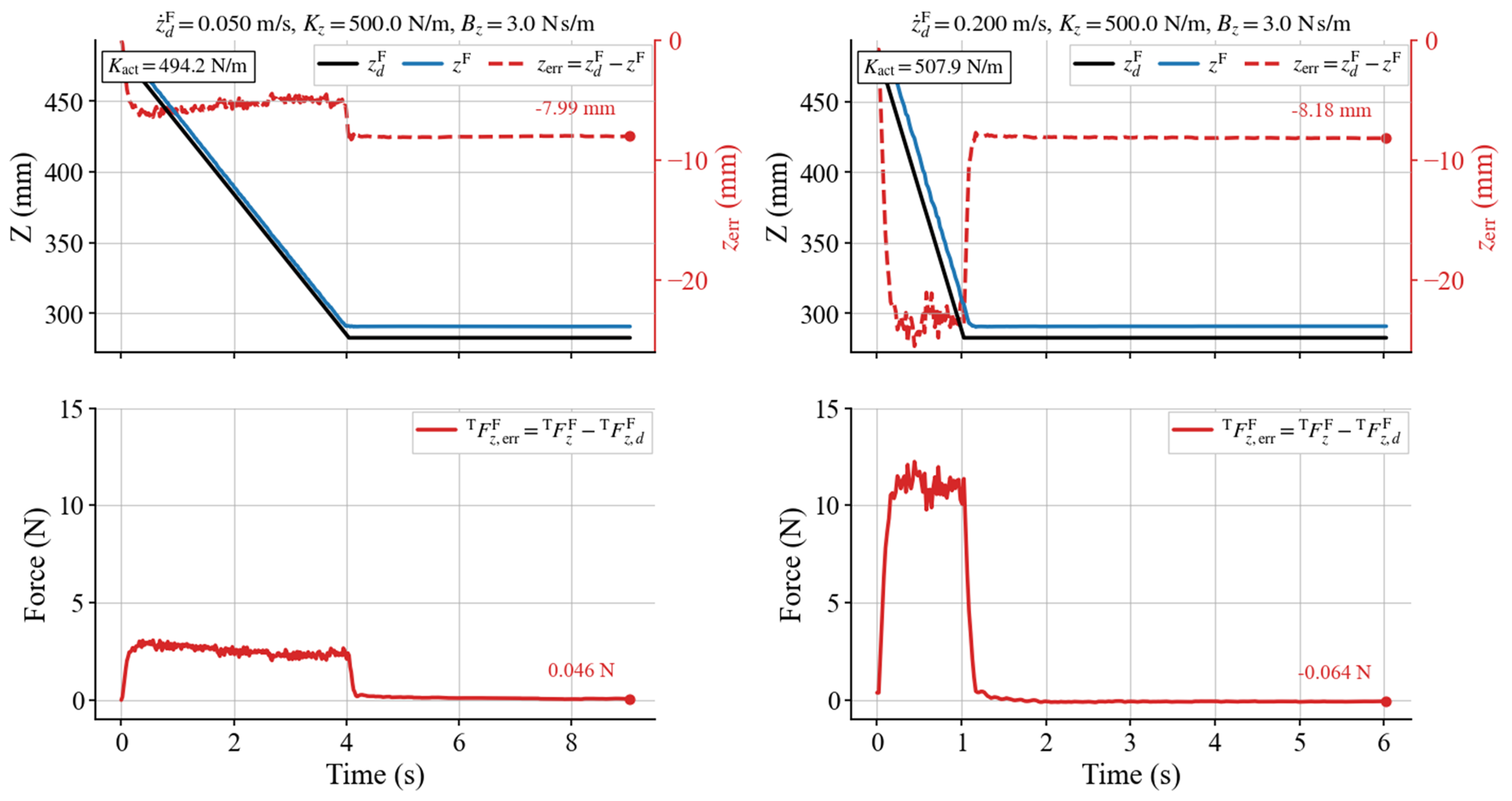}
  \caption{Follower-side collision response under two descent
  velocities, showing the evolution of TCP position error and
  normal force during pre-contact, contact, and steady-state
  phases.}
  \label{fig:coll}
\end{figure}

\noindent\textbf{Collision stability and steady-state
compliance.}
\begin{figure}[t]
  \centering
  \includegraphics[width=0.42\linewidth]{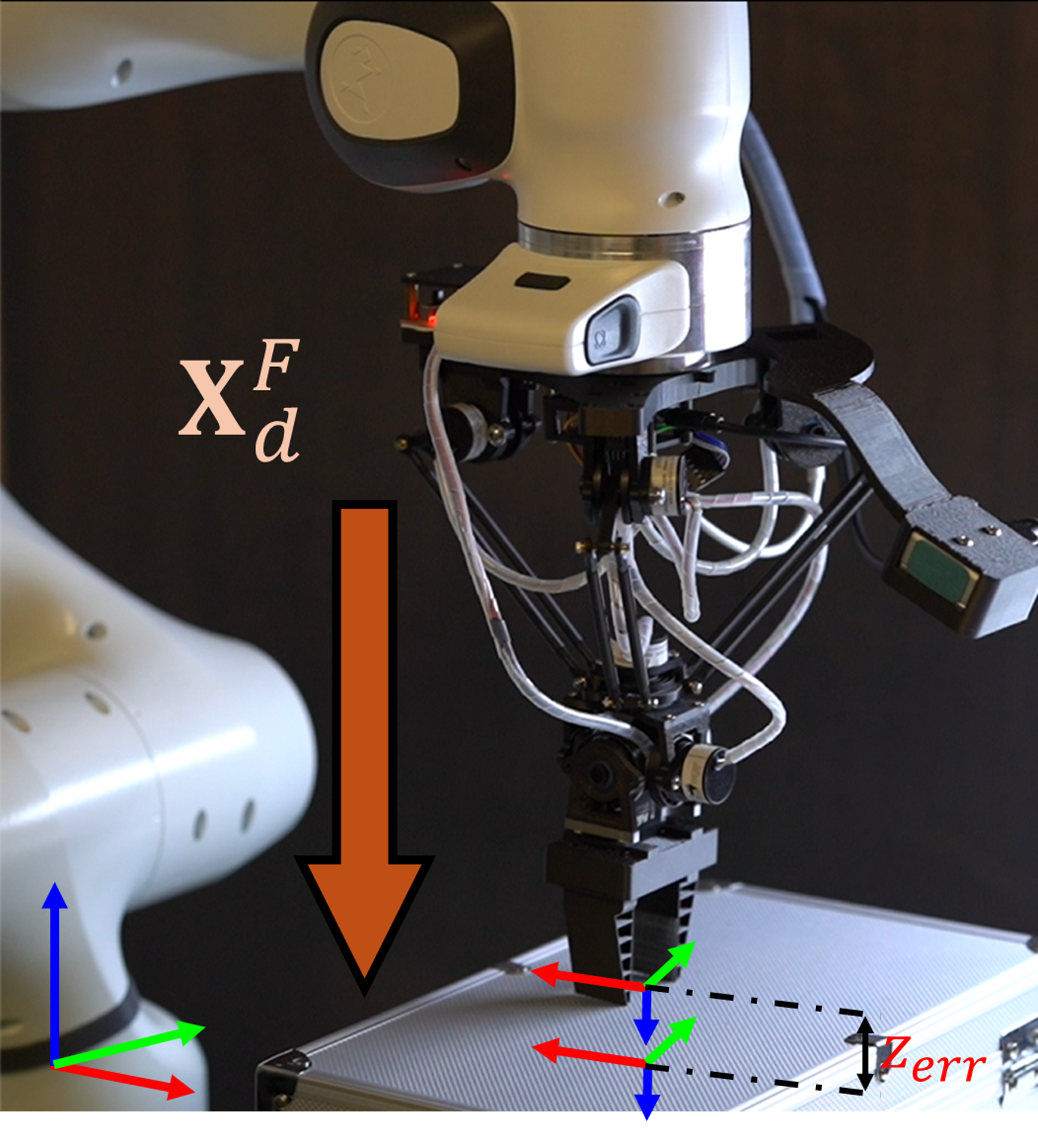}
  \caption{Setup for follower-side collision stability tests: a
  rigid plate is placed 192~mm below $\mathbf{X}^{F}_{0}$; the
  TCP is commanded to a target 200~mm below $\mathbf{X}^{F}_{0}$
  at two velocities.}
  \label{fig:coll_setup}
\end{figure}
A rigid plate is placed 192~mm below $\mathbf{X}^{F}_{0}$ along
the tool-frame $z$ axis (Fig.~\ref{fig:coll_setup}); the desired
TCP pose is commanded 200~mm below $\mathbf{X}^{F}_{0}$ at
constant velocities $\dot z^F_d\in\{0.05,0.2\}$ m/s. Contact
therefore establishes a persistent $\sim$8~mm pose offset, against
which the position error $z_{\mathrm{err}}=z^F_d-z^F$ and the
normal contact force can be related. As shown in
Fig.~\ref{fig:coll}, the pre-contact force plateau matches the
predicted virtual response $F\approx Kz_{\mathrm{err}}+B\dot
z^F_d$. At impact, the normal force error
$F^F_{z,\mathrm{err}}$ converges rapidly without overshoot,
indicating a smooth transition. The ratio of steady-state $F^F_z$
to $z_{\mathrm{err}}$ recovers the prescribed $K=500$~N/m within
measurement uncertainty. Notably, this compliant contact behavior
is achieved at a moderate 150~Hz control rate, which is enabled by
the inherent mechanical compliance of Delta6; rigid-sensor stacks
typically require kHz-rate loops for comparable stability.

\subsection{Bilateral Network-Emulation Tests}
\begin{figure}[t]
  \centering
  \includegraphics[width=\linewidth]{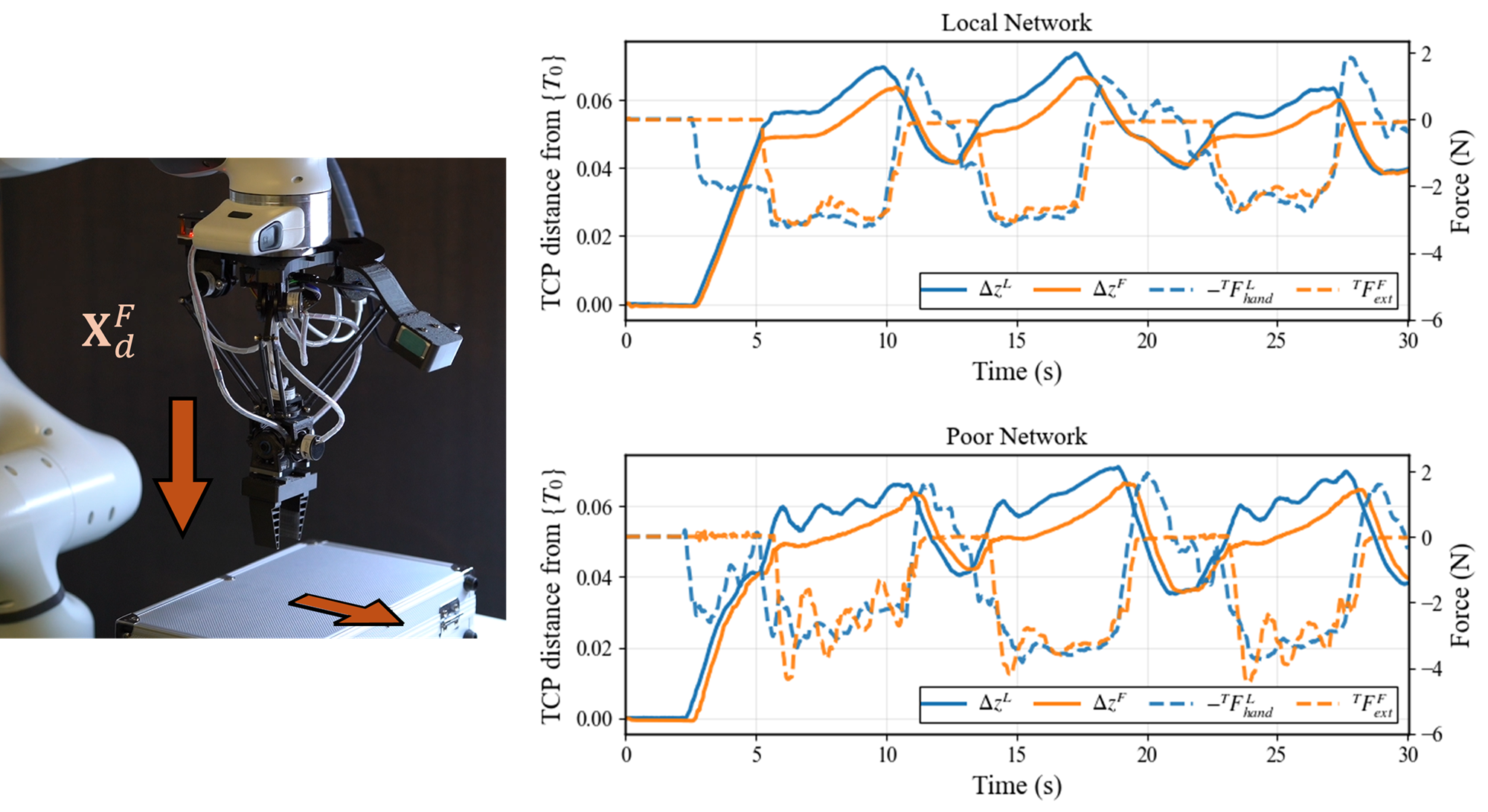}
  \caption{Passivity-oriented bilateral teleoperation experiment
  under contact interaction. The operator drives the follower to
  $\sim$3~N normal contact on a rigid plate, drags it laterally,
  and lifts off, repeated three times.}
  \label{fig:passive}
\end{figure}

To assess stability and operability of the full bilateral pipeline,
the operator uses the leader to drive the follower into $\sim$3~N
normal contact on a rigid plate, drags laterally, and lifts off
(Fig.~\ref{fig:passive}, left). The motion is repeated three
times. Relative displacements $\Delta z^F$, $\Delta z^L$, the
leader hand force $-\prescript{T}{}{F}^L_{z,\mathrm{hand}}$, and
the follower environment force
$\prescript{T}{}{F}^F_{z,\mathrm{ext}}$ are logged.

Under the \emph{Local} condition, $\Delta z^F$ and $\Delta z^L$
follow the same trend except during the brief contact
establishment and release transients, where $\Delta z^L$ continues
to determine the follower setpoint through \eqref{eq:xFd} while
$\Delta z^F$ is constrained by contact. After stable contact,
$\prescript{T}{}{F}^F_{z,\mathrm{ext}}$ and
$-\prescript{T}{}{F}^L_{z,\mathrm{hand}}$ remain near $-3\,$N with
close correspondence, indicating consistent bilateral force
transmission (Fig.~\ref{fig:passive}, right). Under the
\emph{Poor} condition a small free-space offset between
$\Delta z^F$ and $\Delta z^L$ appears due to delayed and jittered
setpoint updates, and mild oscillations appear in $\Delta z^L$
and $\prescript{T}{}{F}^F_{z,\mathrm{ext}}$ during sliding contact.
These oscillations do not grow over time and can be regarded as
operating near the stability boundary; under \emph{Good}/\emph{Fair}
they are not noticeable.

\begin{figure}[t]
  \centering
  \includegraphics[width=\linewidth]{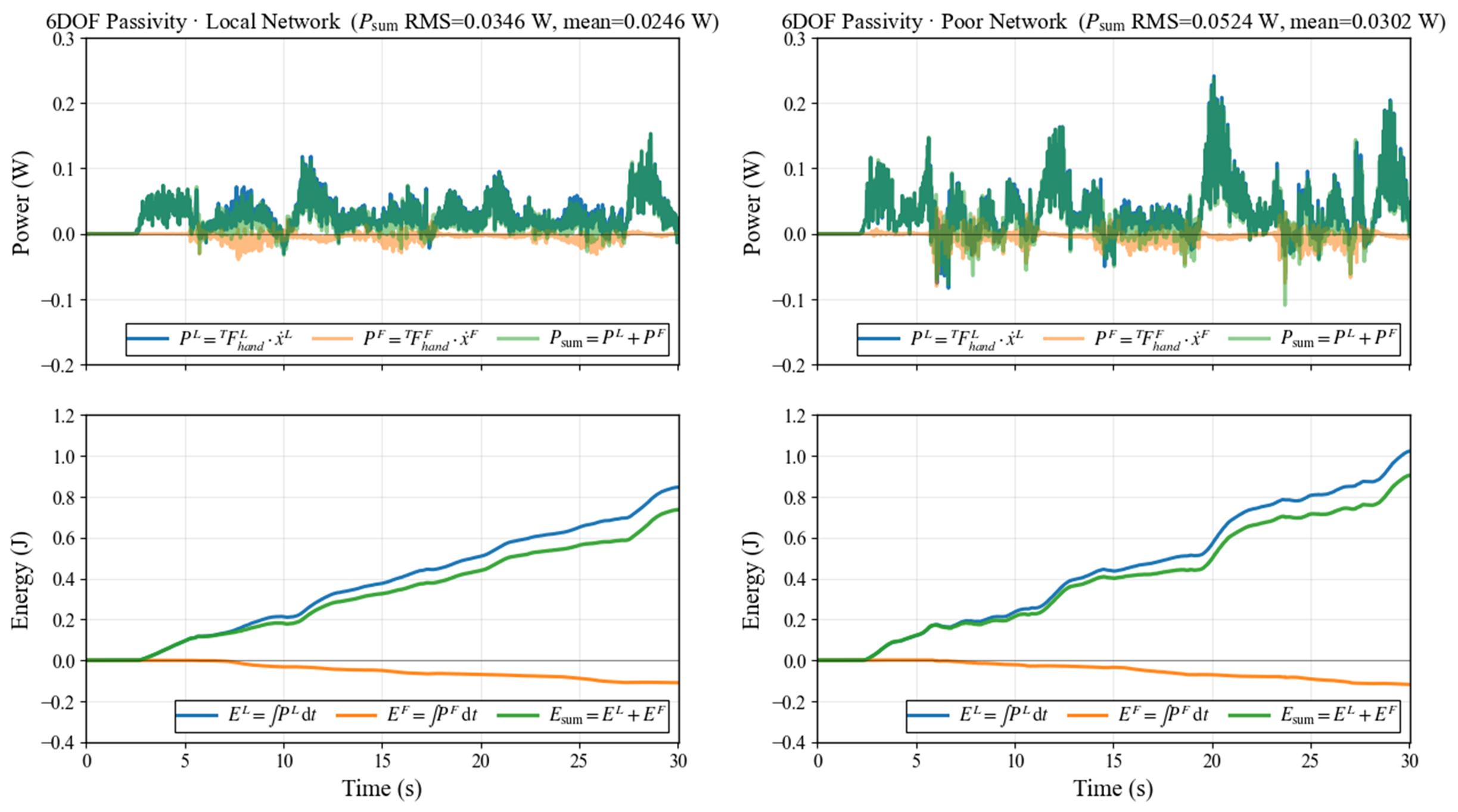}
  \caption{Power and energy analysis under \emph{Local} and
  \emph{Poor} network conditions. The cumulative energy
  $E_{\mathrm{sum}}$ remains positive throughout, indicating that
  the injected operator energy consistently exceeds the extracted
  environment energy.}
  \label{fig:passive_plot}
\end{figure}

Fig.~\ref{fig:passive_plot} reports the net power $P_{\mathrm{sum}}$
and cumulative net energy
$E_{\mathrm{sum}}=\int P_{\mathrm{sum}}\,dt$. Under both network
conditions the operator-input work exceeds the
environment-output work throughout the interaction, primarily due
to the relatively large leader-side admittance damping
$\mathbf{B}^L$ chosen for dissipation. The cumulative balance
$E_{\mathrm{sum}}$ remains positive over the entire trial,
indicating that the system does not generate energy spontaneously.
The average $P_{\mathrm{sum}}$ is slightly lower under
\emph{Local} than under \emph{Poor}, reflecting the additional
operator work required to overcome the small jitter and asynchrony
introduced by delay and packet loss.

\section{Discussion and Limitations}
The proposed architecture closes the loop at a moderate 150~Hz
without sacrificing contact stability, primarily because the
Delta6 end-effector adds mechanical compliance between the
manipulator flange and the environment. This relaxes the typical
kHz-rate requirement reported for rigid F/T sensor
chains~\cite{fang2023rh20t,Panzirsch2022ScienceRobotics} and
shifts much of the design effort to the middleware time/rate
contract and the choice of damping/notch parameters. The
three-rate decoupling and the unified WOS Cartesian interface
make heterogeneous leader/follower combinations operationally
trivial: in our experiments a UFactory Lite6 leader drives an FR3
follower without modifying the application logic.

Two limitations remain. First, leader-side responsiveness is
constrained by the bandwidth of the underlying motion component;
under the chosen damping, the closed-loop $-3\,$dB cutoff is in
the low single-Hz range, which is sufficient for polishing or
compliant assembly but not for fast free-space teleoperation.
Second, although the energy signature is favorable across the
tested conditions, the controller does not include an explicit
passivity-preserving mechanism; under more adverse networks or
stiffer contact, the asynchrony observed in the \emph{Poor}
condition could grow.

\section{Conclusion}
We have presented a bilateral teleoperation framework that
combines a hardware-agnostic Cartesian middleware (WOS) with a
low-cost compliant 6-DOF end-effector force sensor (Delta6) and a
matched pair of damping-only Cartesian admittance / stiffness--
damping impedance loops, augmented by a 6-D biquad notch filter
on the leader side. On a Lite6/FR3 testbed, the system maintains
stable closed-loop tracking under emulated wide-area network
conditions up to $120\pm40$~ms delay with 1\% packet loss, matches
the prescribed steady-state contact compliance at 150~Hz, and
exhibits a favorable cumulative energy signature in
passivity-style tests. Future work will pursue higher-bandwidth
leader-side motion hardware, explicit passivity-preserving
bilateral schemes for adverse network conditions, and integration
with downstream visuo-haptic LfD pipelines.

\bibliographystyle{IEEEtran}

\end{document}